# A Unified Method for First and Third Person Action Recognition

Ali Javidani
Department of Computer Science and Engineering
Shahid Beheshti University
Tehran, Iran
a.javidani@mail.sbu.ac.ir

Ahmad Mahmoudi-Aznaveh
Cyberspace Research Center
Shahid Beheshti University
Tehran, Iran
a_mahmoudi@sbu.ac.ir

*Abstract*—In this paper, a new video classification methodology is proposed which can be applied in both first and third person videos. The main idea behind the proposed strategy is to capture complementary information of appearance and motion efficiently by performing two independent streams on the videos. The first stream is aimed to capture long-term motions from shorter ones by keeping track of how elements in optical flow images have changed over time. Optical flow images are described by pre-trained networks that have been trained on large scale image datasets. A set of multi-channel time series are obtained by aligning descriptions beside each other. For extracting motion features from these time series, PoT representation method plus a novel pooling operator is followed due to several advantages. The second stream is accomplished to extract appearance features which are vital in the case of video classification. The proposed method has been evaluated on both first and third-person datasets and results present that the proposed methodology reaches the state of the art successfully.

## I. Introduction

Video recognition is one of the popular fields in artificial intelligence that aims to detect and recognize ongoing events from videos. This can help humans to inject vision to robots in order to assist them in different situations. For instance, one of the most prominent applications of video classification is driver-less cars which are going to become available in the market.

Totally, there are two main categories of videos that researchers conduct their experiments on them: third-person and first-person videos. In third-person videos, most of the times, camera is located in a specific place without any movement or scarcely with a slight movement and records the actions of humans [1]; while, in first-person videos the person wears the camera and involves directly in events [2]. This is the reason that first-person videos are full of ego-motion and recognizing activities in them is highly challenging [3]. Due to the fact that circumstances of camera and recording actions are completely different from each other in the first and third person videos, there exist two main approaches for classifying each group and to the best of our knowledge, there does not exist a unified method that works perfectly for both.

The main motivation of our work is to provide a unified framework which can classify both first and third-person videos. Toward this goal, two complementary streams are designed to capture motion and appearance features of video data. The motion stream is based on calculating optical flow images to estimate short-term motion in video and by following them over time, using PoT representation method with different pooling operators, long-term motion dynamics are extracted efficiently. The appearance stream is obtained via describing the middle frame of the input video utilizing pre-trained networks. Our method is evaluated on two different datasets UCF11 (third-person) [1] and DogCentric (first-person) [2]. It is demonstrated that the proposed method achieves high accuracy for both datasets.

## II. Related Works

In general, there are two major approaches for classifying videos: Traditional and Modern approaches. Traditional ones are based on descriptors which try to detect different aspects of each action. At the first step, features of video segments are extracted. These features can be interest points or dense points obtained from raw input frames [4, 5]. Harris3D is one of the ways to obtain 3D corner points from video [6]. Then feature points are described by handcrafted descriptors such as HOG [7], HOF [8] and MBH [9]. To describe features more effectively, some of these descriptors have been extended to 3 dimensions to incorporate temporal information in their calculations. HOG3D [10] and SIFT3D [11] are two of the most popular ones.

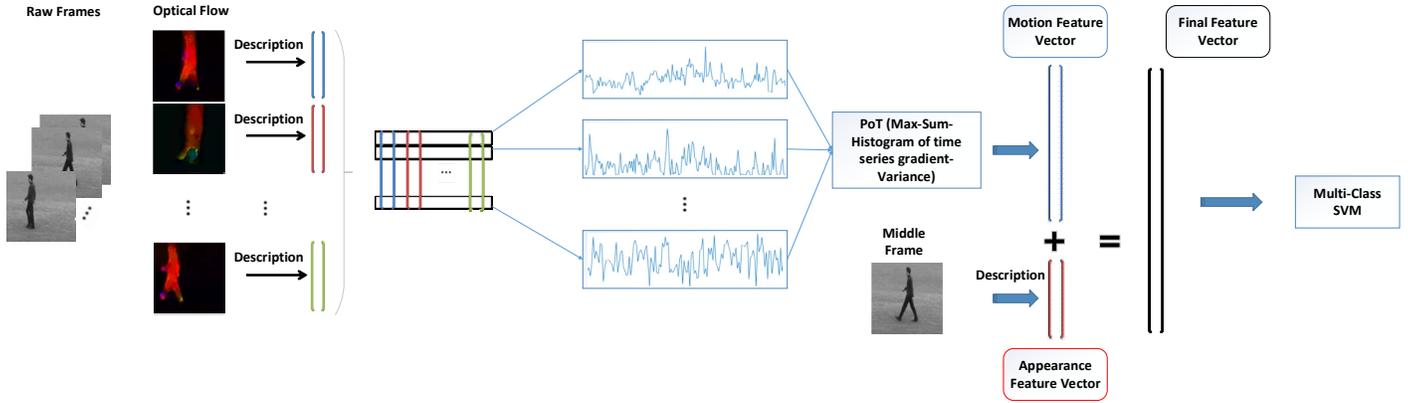

Figure 1: General pipeline of the proposed methodology. Our framework has two streams for obtaining motion and appearance features. The top stream extracts motion, while the bottom extracts appearance features.

Following feature extraction and description phases, in order to obtain a constant-length feature vector and becoming independent of some variables, such as number of frames for each video or number of interest points, an encoding step is required. For so doing, encoding methods like Bag of Visual Words (BoVW) [12], Fisher kernel [13] have been used till now. The experiments illustrate that fisher kernel is more accurate than the former one. However, recently PoT encoding method was proposed by Ryoo *et al.* and it could reach state-of-the-art results in the case of first-person videos [3].

Modern approaches are mostly based on deep learning. Convolutional Neural Networks (CNNs) could succeed in giving the best results on image recognition, image segmentation and so forth [14-16]. Although one problem in the case of video domain is that these networks are designed for two-dimensional input images. To address this problem, some researches have been conducted. As a case in point, Karpathy *et al.* introduced four models of two-dimensional but multi-channel CNNs and in their models time dimension was incorporated in different channels of the network [17]. Zisserman *et al.* proposed two stream CNN model to classify videos. However, their method suffers from the problem that the number of stacked optical flow frames given to CNN is limited due to the problem of overfitting of the network [18].

Furthermore for better estimation of motions in video, a 3-dimensional convolutional neural network (C3D) was devised [19]. All convolution and pooling layers in C3D operate 3D and the depth of time dimension convolution is a small number due to the vast amount of convolution calculations. Hence, it can only capture the short-term motion dynamics and longer ones would be lost by using this network. A recent work used stacked auto-encoders to obtain tractable dimensionality for the case of first-person videos [20]. Also, another work designed a deep fusion framework in which with the aid of LSTMs, representative temporal information have been extracted and could reach state-of-the-art results on three widely used datasets [21].

## III. PROPOSED METHOD

In this section, we introduce our proposed method pipeline. Generally, videos either third-person or first-person, consist of two different aspects: appearance and motion. Appearance is related to detecting and recognizing existing objects in each frame, while motion is their following over time. Hence, motion information is highly correlated with temporal dimension. As it is depicted in Fig. 1, in order to capture two mentioned aspects in video, our proposed framework has two streams independent of each other. The upper stream is for extracting motion and the bottom is for appearance features.

In the following, motion feature extraction is explained in more detail. Firstly the images of optical flow between consecutive frames are calculated. This helps to estimate short-term motion through nearby frames for each video. However, estimating long-term motion is strictly challenging and is still an open area of research.

Here, the idea is to keep track of how short-term motion elements vary over time to estimate longer changes. Therefore, optical flow images should be described by a specific set of features to be pursued over the time dimension. We found that the best way for doing so is utilizing pre-trained networks that already have been trained on large scale image datasets. By doing this, not only training of deep networks, which is a highly time consuming process, is not needed but also a strong representation would be obtained since these networks could reach state-of-the-art results for image recognition.

Aligning representation of sequential frames beside each other leads to obtaining a set of multi-channel time series. There are various ways to extract features from time series. For doing so, PoT representation plus a novel pooling operator is chosen due to several prominent reasons. Firstly, thanks to the temporal filters, time series are break down to sub-intervals which assists to represent high-level activity from lower levels. Furthermore, PoT is benefited from extracting different features from time series that each of them can represent different aspects of data. The resulted time series, especially those coming from first person videos which are full of ego-motion, are more sophisticated than a specific feature (e.g. max) can represent them. As a result, PoT framework can be beneficial for extracting motion features from time series.

PoT representation method extracts different features *max, sum and histogram of time series gradient* from time series. The final feature vector representation, which is designed to be motion features in our framework, is the concatenation of all of these features together for each time series. We add *variance* as another pooling operator to the pooling set and demonstrate that this feature can also extracts useful information. Below is the definition of each pooling operator in the time domain $[t^s, t^e]$. $f_i(t)$ is the value of $i^{th}$ time series in time $t$. The *max* and *sum* pooling operators are defined as:

$$p_i^{max}[t^s, t^e] = \max_{t=t_s \ldots t_e} f_i(t) \qquad (1)$$

$$p_i^{sum}[t^s, t^e] = \sum_{t=t_s}^{t_e} f_i(t) \qquad (2)$$

and the histograms of time series gradient pooling operators are defined as:

$$p_i^{\Delta^+}[t^s, t^e] = \sum_{t=t_s}^{t_e} h_i^+(t), \quad p_i^{\Delta^-}[t^s, t^e] = \sum_{t=t_s}^{t_e} h_i^-(t) \qquad (3)$$

where

$$h_i^+(t) = \begin{cases} f_i(t) - f_i(t-1) & if\ (f_i(t) - f_i(t-1)) > 0 \\ 0 & otherwise \end{cases} \qquad (4)$$

$$h_i^-(t) = \begin{cases} f_i(t-1) - f_i(t) & if\ (f_i(t) - f_i(t-1)) < 0 \\ 0 & otherwise \end{cases} \qquad (5)$$

We proposed *variance* as a new pooling operator as follows:

$$p_i^{var}[t^s, t^e] = \sum_{t=t^s}^{t^e} \frac{(f_i(t) - \bar{f})^2}{(t^s - t^e + 1)}, \qquad (6)$$

$$\bar{f} = \frac{\sum_{t=t^s}^{t^e} f_i(t)}{(t^s - t^e + 1)} \qquad (7)$$

Besides, in order to concentrate better on the resulted time series, temporal pyramid filters are applied on each series. Hence, the resultant time series are: the whole time domain, half-parts of the whole time domain, one-fourth parts and so forth.

Motion features are extracted as explained above by applying pooling operators on each level of the resulted time series after exploiting temporal pyramid filters. On the other hand, appearance features have substantial role in classifying videos. In our pipeline, we used middle frame of each video and feed it to a pre-trained network to obtain appearance feature vector. The final representation for each video is acquired from concatenating motion and appearance feature vectors that are expected to represent complementary information. As the last step, a SVM is trained on the final feature vector.

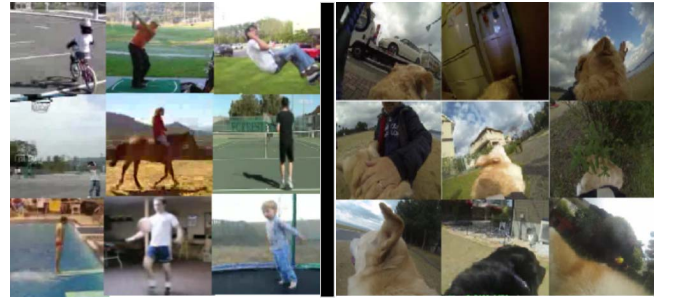

Figure 2: Some sample frames from different classes of two different datasets UCF11 (left) and DogCentric (right).

## IV. EXPERIMENTAL RESULTS

We conducted our experiments on two public datasets: UCF11 and DogCentric. Fig. 2 represents some sample frames of them. UCF11 is a third-person type which contains 11 different human activities e.g. playing basketball, volleyball and it consists of about 1600 human activity videos. In this dataset camera is not usually located in a specific place and it has large amounts of movement. To evaluate our method, we performed Leave One Out Cross Validation (LOOCV) on this dataset as in the original work [1].

DogCentric is a first-person activity dataset in which a camera is located on the back of dogs. Thus, it has large amounts of ego-motion which makes it highly challenging [2]. It has 10 class activities consisting activities of dogs such as walking and drinking as well as interacting with humans. The total number of videos in this dataset is 209. Like other previous methods, half of the total number of videos per class is used as training and the other half is used as testing. For classes with odd number of clips, the number of test instances is one more than number of training. We performed our algorithm 100 times with different permutation for train and test sets. The mean classification accuracy is reported in Table 1. It is clear that the proposed method has been achieved a significant improvement in terms of classification accuracy in compare with two traditional representation methods Bag of Visual Words (BoVW) and Improved Fisher Vector (IFV). In addition, the proposed method could outperform the baseline PoT method in most classes of the DogCentric dataset and also in the final accuracy.

TABLE 1: COMPARISON OF THE ENCODING METHODS ON THE DOGCENTRIC DATASET. F1-SCORES PER CLASS AND FINAL CLASSIFICATION ACCURACY.

| Activity Class | Method Accuracy (%) | | | |
|---|---|---|---|---|
| | BoVW | IFV | PoT | **Proposed** |
| Ball Play | 48.29 | 61.46 | 74.66 | **74.73** |
| Car | 73.23 | 81.54 | 89.54 | **89.60** |
| Drink | 12.60 | 45.23 | 64.02 | **64.34** |
| Feed | 50.62 | 57.64 | **59.41** | 57.35 |
| Turn Head(left) | 29.73 | 31.13 | 29.15 | **32.96** |
| Turn Head(right) | **50.44** | 31.82 | 35.57 | 36.85 |
| Pet | 42.85 | 67.65 | **78.34** | 77.23 |
| Body Shake | 41.33 | 46.54 | 72.70 | **72.76** |
| Sniff | 54.93 | 64.23 | 75.51 | **76.10** |
| Walk | 44.69 | 44.91 | 56.42 | **56.87** |
| Final accuracy | 47.39 | 55.68 | 65.00 | **67.12** |

For obtaining optical flow images of consecutive frames, we used popular HORN & SCHUNCK method [22]. To convert it to colorful images, in order to be fed to pre-trained networks, flow visualization code of Baker *et al.* [23] was followed. In our implementation, this method was applied to all frames of each video and we did not sample the existing frames. GoogLeNet, as a pre-trained network, is utilized to describe either optical flow images (in motion stream) or middle frame (in appearance stream). This was feasible by omitting softmax and fully-connected layers of the network. GoogLeNet has 1024 neurons in this layer.

Furthermore in the case of number of temporal pyramids, different experiments have been conducted and results are illustrated in the Table 2. It can be seen that by increasing the number of temporal pyramids up to 4 levels, the classification accuracy has been improved. While by increasing it to five levels, it has decreased in compare with four levels. We believe this phenomenon is due to the fact that by increasing the temporal pyramid levels, the number of dimensionality would increase dramatically. On the other hand, there do not exist enough training data for learning the classifier. This is the reason that increasing number of temporal pyramids cannot always improve the performance of the system.

The proposed method is also evaluated on third-person video dataset UCF11. The number of temporal pyramids used for this dataset is 3 and sampling between frames was not performed. Comparison of our method to the state-of-the-art results on this dataset is reported in Table 3. As can be seen, the proposed method with chi-squared SVM could reach the best results on this dataset.

In all our experiments SVM classifier with linear and chi-squared kernel is used and the later one showed better performances.

## V. CONCLUSION

In this paper, a new approach for video classification was proposed which has the capability of employing for two different categories of first and third-person videos. Short-term/long-term motion changes are calculated by extracting discriminant features from motion time series following PoT representation method with a novel pooling operator. Final feature vector is resulted from concatenating two complementary feature vectors of appearance and motion to perform the classification. By evaluating the proposed method on two different types of datasets and comparing the obtained results to the state of the art, it is concluded that the proposed method not only works perfectly for both groups but also increases the accuracy.

TABLE 2: COMPARISON OF CLASSIFICATION ACCURACY OF THE DOGCENTRIC DATASET ACCORDING TO TEMPORAL PYRAMID LEVELS.

| Number of temporal pyramids | Classification accuracy |
|---|---|
| 1 | 58.25% |
| 2 | 59.66% |
| 3 | 66.33% |
| 4 | 67.12% |
| 5 | 66.85% |

TABLE 3: COMPARISON OF OUR RESULTS TO THE STATE-OF-THE-ART APPROACHES ON UCF11 DATASET.

| Method | Accuracy |
|---|---|
| Hasan et al. [24] | 54.5% |
| Liu et al. [1] | 71.2% |
| Ikizler-Cinbis et al. [25] | 75.2% |
| Dense Trajectories[26] | 84.2% |
| Soft attention[27] | 84.9% |
| Cho et al. [28] | 88.0% |
| Snippets[29] | 89.5% |
| Two Stream LSTM(conv-L) [21] | 89.2% |
| Two Stream LSTM(fc-L) [21] | 93.7% |
| **Proposed method(Linear SVM)** | **93.75%** |
| Two Stream LSTM(fu1) [21] | 94.2% |
| Two Stream LSTM(fu2) [21] | 94.6% |
| **Proposed method(chi-squared SVM)** | **95.73%** |